\documentclass[10pt,a4paper,twoside]{article}
\usepackage{JISE}
\usepackage{graphicx}
\usepackage{times}
\usepackage{cite}
\usepackage{mathptmx}
\usepackage{amsmath,amssymb,amsfonts}
\usepackage{algorithmic}
\usepackage{subcaption}
\usepackage{booktabs}
\usepackage{bbm}
\usepackage{amsthm}
\usepackage{hyperref}
\usepackage{algorithm}
\usepackage{multirow}
\def\BibTeX{{\rm B\kern-.05em{\sc i\kern-.025em b}\kern-.08em
    T\kern-.1667em\lower.7ex\hbox{E}\kern-.125emX}}

\newcommand\myeq{\mathrel{\overset{\makebox[0pt]{\mbox{\normalfont\tiny\sffamily def}}}{=}}}
\theoremstyle{definition}

\newtheorem{theorem}{Theorem}
\newtheorem{lemma}{Lemma}

\title{Efficient Genotype Classification from 3D MRI in ADPKD via Density-Aware Point Sampling and E(3)-invariance}
\author{Kwanghee Lee{\small$^1$}, Hyeongsub Lee{\small$^1$}, \\ Youngwoo Kim{\small$^{1,+}$}, and Hyunji Jeong{\small$^{2,+}$}\\
  {\small\em {\small$^1$} School of Computer Engineering, Kumoh National Institute of Technology} \\
  {\small\em Gumi, Republic of Korea}\\
  {\small\em {\small$^2$} School of Artificial Intelligence, Kongju National University} \\
  {\small\em Cheonan, Republic of Korea}\\
  {\small\em E-mail: \{kwanghee; mesubby; ywkim09\}@kumoh.ac.kr;}\\
  {\small\em hjjeong@kongju.ac.kr}
}

\authorrunning{K. Lee, H. Lee, Y. Kim, and H. Jeong.}
\titlerunning{E3PN}

\begin{document}
\editorfootnote{Received March 11, 2014; revised June 20, 2014; accepted September 28, 2014. \\
{\small$^+$}Corresponding author}
\maketitle

\begin{abstract}
Early detection of genetic kidney diseases, especially autosomal dominant polycystic kidney disease (ADPKD), is crucial because certain genotypes such as truncating PKD1 mutations cause rapid kidney growth and early progression to end-stage kidney disease.
MRI-based genotype inference is a potential low-cost complement to genetic testing. However, although many deep learning methods have been developed for classifying various MR images, applying them to ADPKD remains challenging: (1) abdominal MR images are generally too large to be handled by deep learning models designed for smaller images like brain MRI, and (2) the amount of labeled data for ADPKD genetic mutations is insufficient to train deep learning models. 
In order to address these challenges, we propose E3PN, an efficient 3D MRI-based ADPKD genotype classification network that relies on point sampling and E(3)-invariance. We utilize a point representation of 3D MRI, which consumes less GPU memory than a voxel representation. For further memory efficiency, 3D points are sampled from a mixture distribution over the two kidneys in each image. Since E(3)-invariance has been shown to improve data efficiency, we also incorporate it into our model through coordinate canonicalization. 
On an ADPKD dataset of 441 patients, E3PN trains $163\times$ faster and uses $103\times$ less GPU memory than a 3D ResNet, while also improving F1 from 0.165 to 0.226 over both baselines.
These results establish a first benchmark for five-way ADPKD genotype classification from 3D MRI and show that point sampling and coordinate canonicalization yield two-orders-of-magnitude efficiency gains with a modest but consistent improvement in F1. At the same time, all three methods, including a 3D CNN trained on cropped volumes, remain within a few points of chance-level F1 ($\sim$0.2), indicating that genotype is only weakly encoded in kidney morphology at this cohort size. We therefore position E3PN as an efficient, reproducible baseline for this task rather than as a clinically deployable classifier.
\end{abstract}

\section{Introduction}
\label{sec:introduction}
Diagnosing diseases from magnetic resonance imaging (MRI), without resorting to more invasive or costly examinations, has gained significant attention. Deep learning has been central to this progress, obtaining remarkable results in tasks such as brain tumor detection~\cite{litjens2017survey}, kidney segmentation~\cite{isensee2021nnu}, and cardiac MRI classification~\cite{wang2024screening}. Despite these successes, MRI-based diagnosis of genetic kidney diseases especially autosomal dominant polycystic kidney disease (ADPKD) still needs to be explored further.

\begin{figure*}[!t]
    \centering
    \begin{subfigure}[c]{0.35\textwidth}
        \centering
        \includegraphics[width=\textwidth]{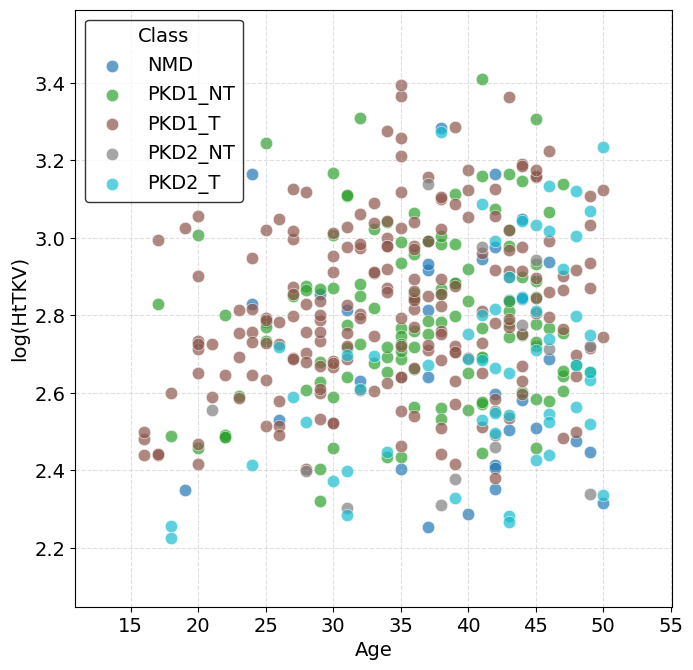}
        \caption{htTKV vs. Age}
        \label{fig:httkv}
    \end{subfigure}
    \hfill
    \begin{subfigure}[c]{0.19\textwidth}
        \centering
        \includegraphics[width=\textwidth]{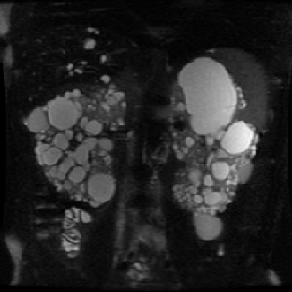}
        \caption{ADPKD}
        \label{fig:adpkd_original}
    \end{subfigure}
    \hfill
    \begin{subfigure}[c]{0.19\textwidth}
        \centering
        \includegraphics[width=\textwidth]{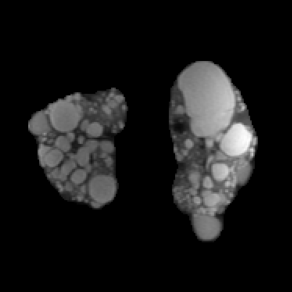}
        \caption{Segmented}
    \end{subfigure}
    \hfill
    \begin{subfigure}[c]{0.19\textwidth}
        \centering
        \includegraphics[width=\textwidth]{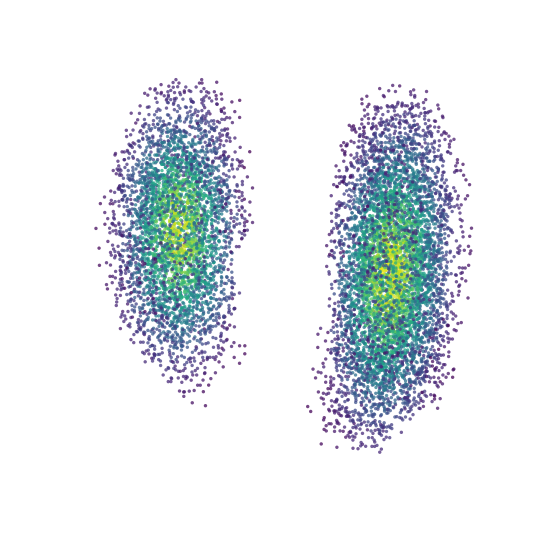}
        \caption{Point set}
    \end{subfigure}

    \begin{subfigure}[b]{\textwidth}
        \centering
        \includegraphics[width=\textwidth]{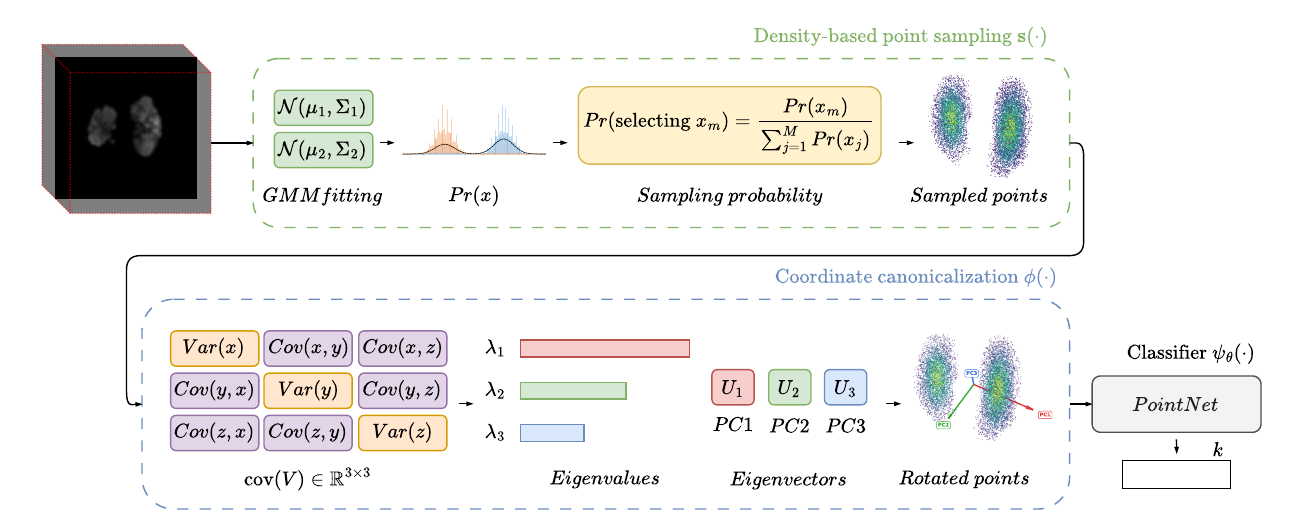}
        \caption{Overview of the E3PN pipeline}
        \label{fig:overview}
    \end{subfigure}

    \caption{(a) A distribution of height-adjusted total kidney volumes (htTKV) across ages for five ADPKD genetic variants. (b)--(d) An ADPKD kidney, its segmentation mask, and a sampled point set. (e) Overview of the E3PN pipeline. }
    \label{fig:data_and_pipeline}
\end{figure*}

ADPKD leads to progressive kidney enlargement and, without treatment, to end-stage kidney disease and premature death, typically by the fifth or sixth decade of life~\cite{halvorson2010polycystic}. Since ADPKD has a genetic cause, genetic testing is a reliable diagnostic, but it is time-consuming and expensive. 
Alternatively, MRI-based approaches can reduce both time and cost. For example, by using a patient's age, height, and kidney volume, one can promptly estimate the state of ADPKD progression~\cite{irazabal2015imaging}. However, height-adjusted total kidney volume (htTKV) is insufficient to capture the progression speed of ADPKD (Figure~\ref{fig:httkv}), which is critical for early intervention to save patients' lives. This observation is consistent with the results in~\cite{kim2019genetic}. Thus, such a method is of limited use for estimating the progression speed.

Deep learning-based approaches are promising for classifying MRIs without hand-crafted features. 
Although convolutional neural networks are key to success, we identify the following issues in applying them to ADPKD classification.
\begin{itemize}
    \item High memory consumption: Vanilla convolutions consume a large amount of GPU memory to process 3D MR images of the human body (e.g., a $64\times256\times256$-sized MR image has 4,194,304 voxels). 
    \item Data scarcity issue: ADPKD datasets are insufficient for training conventional deep learning models for MRI classification.
\end{itemize}

One might consider image cropping to reduce GPU memory consumption, but it is hardly a viable solution because kidney volumes vary between patients, and some kidneys under ADPKD can be nearly as large as the entire abdomen (see Figure~\ref{fig:adpkd_original}), meaning the cropped volume remains nearly as large as the original. 
For the second issue, using public MRI datasets is a possible solution, but they typically focus on brain imaging or normal kidneys.
Thus, a method that efficiently utilizes kidney datasets is needed. 

To this end, we propose E3PN, which addresses both challenges simultaneously---it replaces the dense voxel grid with a lightweight point cloud tailored to kidney geometry, and reduces the orientation variability the classifier must learn from limited training data by canonicalizing the point coordinates.
The proposed method has the additional advantage of enabling efficient training of deep models on a single GPU, even with large datasets. 

\begin{figure}[t]
    \centering
    \includegraphics[width=0.4\textwidth]{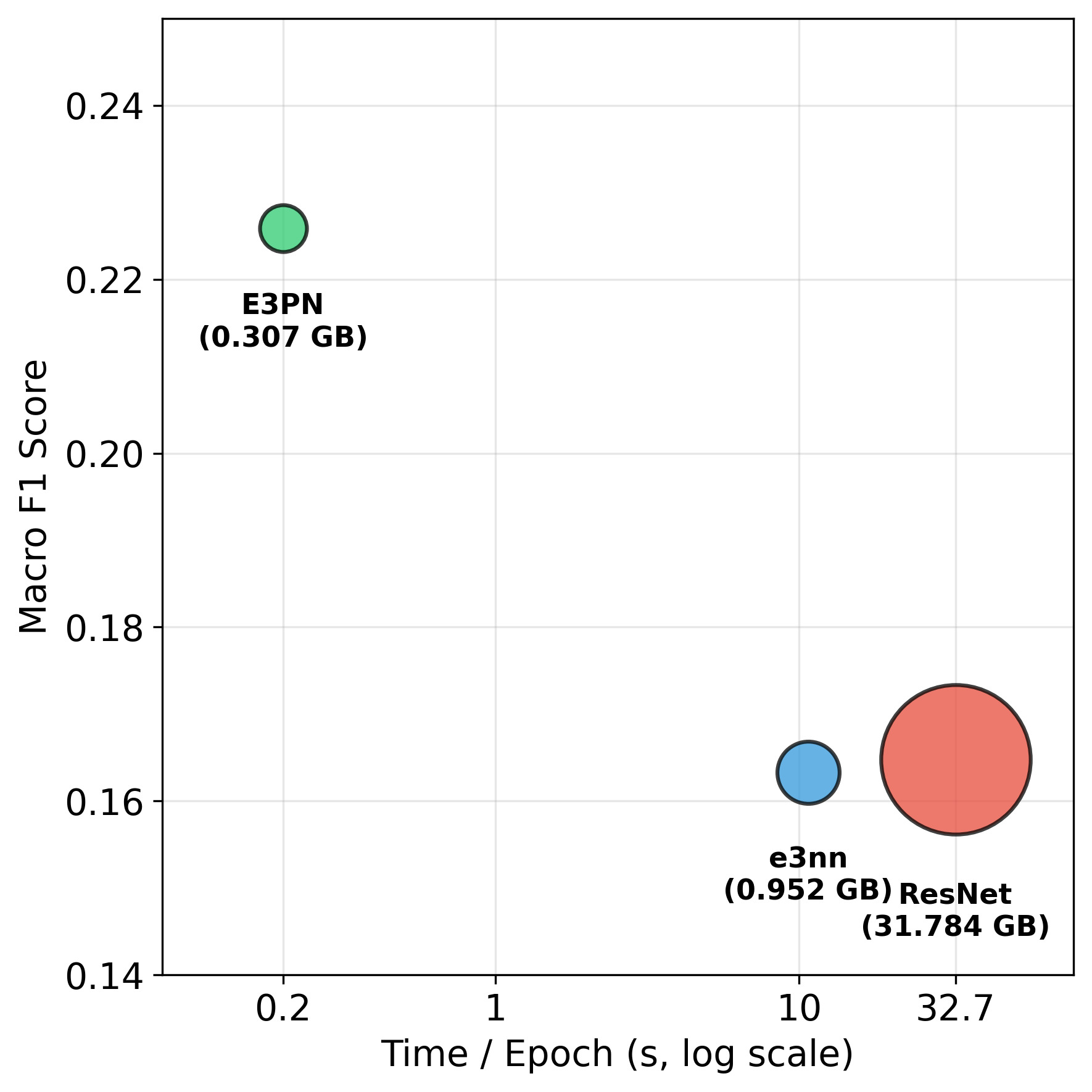}
    \caption{E3PN attains higher F1 than both baselines at roughly two orders of magnitude lower training time and memory.}
    \label{fig:bubble}
\end{figure}

The main contributions of this paper are as follows.
First, we propose E3PN, which combines density-based point sampling with provably E(3)-invariant coordinate canonicalization, enabling single-GPU training on full abdominal MRI volumes at $163\times$ the speed and 1/103 the memory of a 3D CNN baseline. 
Second, to the best of our knowledge, we present the first benchmark for five-way ADPKD genotype classification from 3D MRI, evaluated with 10-fold cross-validation on a cohort of 441 patients (Figure~\ref{fig:bubble}). 
Third, we find that five-way genotype classification from MRI alone is near chance for all methods tested (F1 score 0.16–0.23 against a chance level of 0.20), with E3PN's gain concentrated in separating the two PKD1 subtypes; we analyze this in Section \ref{sec:Comparisonresults}  and argue it reflects the data regime rather than model capacity.

\section{Related work}\label{sec:relatedwork}
\subsection{Automated Imaging Classification of ADPKD}

Deep learning has been applied to kidney images, although most efforts have focused on segmentation and measuring kidney size rather than classification. Kim et al.~\cite{kim2024deep} developed a deep learning method to automatically classify ADPKD patients into two categories. Their method classifies renal morphology to support volume-based assessment, whereas our work aims to predict the rate of disease progression, which is a different and clinically complementary task. Kremer et al.~\cite{kremer2023magnetic} took a different approach by computing statistical features of kidney MRI scans and using them to distinguish between the two main genetic subtypes of ADPKD. They found that the results varied considerably depending on how the images were preprocessed, with classification performance ranging from near chance to moderate accuracy. Outside of imaging, Panda et al.~\cite{panda2023deep} used a neural network on gene expression measurements from mice to detect PKD early, although this line of work relies on laboratory data rather than clinical images. These studies show that while deep learning-based methods have made progress in ADPKD diagnosis, existing imaging approaches work with either 2D image slices or hand-crafted features. None of them represents the kidney as a 3D point cloud, which is the strategy we pursue in this work.

\subsection{Point Cloud and Equivariant Networks in Medical Imaging}

A point cloud is simply a set of 3D coordinates (and optionally associated features such as intensity), offering a lightweight alternative to dense volumetric grids for representing 3D shapes. Point cloud networks, originally developed for tasks such as object recognition in autonomous driving, have recently been adopted in medical imaging. Wang et al.~\cite{wang2026m} proposed M-PointNet, which extends a well-known PointNet-based architecture with additional feature-extraction layers and training supervision signals. Applied to 3D models of brain blood vessels, their method classifies brain diseases and shows that point cloud networks can handle the irregular geometries in medical data even when training samples are limited. 

Celledoni et al.~\cite{celledoni2021equivariant} demonstrated that building geometric symmetries---invariance to rotations and translations---directly into neural network layers leads to better image reconstructions in CT and MRI, while requiring fewer training examples than standard networks. Their finding underscores a principle central to our work: when a model is aware that the answer should not change if the input is rotated or shifted, it can learn more effectively from small datasets.

Canonicalization is an established alternative to building invariance into network layers: instead of constraining every layer, the input is mapped to a canonical pose and an ordinary network is applied afterwards.
Ma et al.~\cite{ma2024canonicalization} give a general analysis of this view and its relation to equivariant architectures; frame averaging~\cite{puny2022frame} and learned canonicalization~\cite{kaba2023equivariance} are two recent examples.
For point clouds specifically, principal-component frames have been used to obtain rotation invariance~\cite{xiao2020pca,li2021closer}, with the well-known caveat that eigenvector signs must be disambiguated. 
We adopt a PCA-based canonicalization rather than an equivariant architecture because, for our task, the cost of invariance matters: abdominal MRI volumes are large and genotype-labeled scans are few, so a linear-time canonicalization that leaves the classifier untouched is preferable to tensor-product layers.

Our method differs from the above works in two ways. First, we apply point cloud networks to 3D MRI classification rather than to surface-based analysis or image reconstruction. Second, instead of redesigning network layers to be invariant, we attain the same invariance at negligible computational cost by canonicalizing.

\section{Methodology}\label{sec:proposedmethod}

\subsection{Structure overview}
Our goal is to learn a mapping $f$ from an abdominal MRI $X \in \mathbb{R}^{H \times W \times D \times 1}$ to a kidney disease label $y\in \{1, ..., C\}$, such that $\Pr(y|X)=f(X)$. 
We model $f$ as a composition of three functions:
\begin{equation}
    f =\psi_{\theta} \circ \phi \circ \mathbf{s},
\end{equation}
where $\mathbf{s}(\cdot)$ is a sampler, $\phi(\cdot)$ is a canonicalization, and $\psi_{\theta}(\cdot)$ is a classifier. Each component is motivated by a specific challenge, and together they form a pipeline illustrated in Figure \ref{fig:overview}.

\subsection{Point representation}
In this work, we use a 3D point representation to reduce GPU memory consumption. A typical abdominal MR image in our dataset contains roughly four million voxels, making standard 3D convolutions prohibitively expensive on a single GPU.
We observe, however, that after segmentation, only a fraction of these voxels belong to the kidney; the rest are background with zero intensity.
A dense voxel grid must nevertheless store these zero-intensity voxels and hence wastes both memory and computation.
An obvious remedy would be to crop the volume to a tight bounding box around the kidney.
However, kidneys with the truncating PKD1 genotype are severely enlarged---often approaching the extent of the original field of view---so cropping removes only a marginal portion of the volume. 
More importantly, even within a tight bounding box, the background voxels scattered between and around the irregular kidney lobes remain, since a rectangular crop cannot conform to the organ's shape. 
This motivates representing each kidney image not as a dense volume but as a set of $N$ points 
\begin{equation}
    \{p_i \in \mathbb{R}^{3+1} | i=1, ..., N\}
\end{equation}
 (3 for the coordinate and 1 for the intensity), with $N \ll H \times W \times D$, discarding the empty space entirely.


\subsection{Density-based point sampling}\label{sec:sampling}
A second consideration is point sampling to further reduce memory consumption. While any subsampling reduces cost, uniform random sampling treats all voxels equally and can over-represent noisy boundary regions or under-represent the kidney core. Because each patient's segmentation contains two separately distributed kidneys that may differ substantially in size, we fit a mixture of two Gaussians to the non-zero voxel coordinates and sample proportionally to the estimated density. This produces a point set 
\begin{equation}
   \mathbf{s}(X) =  \{p_1, \ldots, p_N\} 
\end{equation}
that represents the kidney's spatial structure.

Specifically, for an MR image $X$, we first collect the $M$ voxels $x_1, \ldots, x_M$ with non-zero intensity. Using only the 3D coordinate of each voxel $x_m$ (ignoring intensity), we fit a Gaussian mixture:
\begin{equation}
    \Pr(x)= \sum_{i}\alpha_i \mathcal{N}(x|\mu_i,\Sigma_i),
\end{equation}
where $\mu_i \in \mathbb{R}^{3}$ and $\Sigma_i \in \mathbb{R}^{3\times3}$ are the component mean and the covariance matrix, respectively, and $\alpha_i \in [0,1]$ is the mixing coefficient with $\sum_{i}\alpha_i=1$. We use two components to match the anatomy: the left and right kidneys form two distinct spatial clusters. 
 
Each voxel $x_m$ is then assigned a sampling probability proportional to its density:
\begin{equation}
   \Pr(\text{selecting } x_m) = \frac{\Pr(x_m)}{\sum_{j=1}^M \Pr(x_j)}.
\end{equation}
We draw $N$ points $p_1, \ldots, p_N \in \mathbb{R}^{3+1}$ according to these probabilities without replacement.
 
\subsection{Coordinate canonicalization}\label{sec:canonicalization}

The second challenge from the Introduction is data scarcity. ADPKD datasets are small, and kidney orientation varies across patients and scanning protocols. Rather than augmenting the data with random transformations, we remove the orientation degree of freedom altogether: we compute a canonical coordinate frame from the eigenvectors of the point set's covariance matrix and re-express all points in that frame. The canonicalized point set $\phi(\mathbf{s}(X))$ is invariant to 3D rotation, translation, and reflection---the Euclidean group E(3)---so the classifier $\psi_\theta$ operates on a fixed canonical frame, independent of the kidney's original orientation.

Let $v_n \in \mathbb{R}^3$ denote the coordinate part of each sampled point $p_n$, and let $V \in \mathbb{R}^{N\times 3}$ be the matrix stacking all $v_n$'s. We compute a canonical coordinate frame via the covariance matrix:
\begin{equation}
    \text{cov}(V) = (V-\mathbf{1}\bar{v}^\top)^{\top}(V-\mathbf{1}\bar{v}^\top) \in \mathbb{R}^{3 \times 3},
\end{equation}
where $\bar{v}=\frac{1}{N}\sum_{n=1}^N v_n$ is the centroid and $\mathbf{1} \in \mathbb{R}^{N}$ is the all-ones vector. Let $U \in \mathbb{R}^{3 \times 3}$ be the orthogonal eigenvector matrix of $\text{cov}(V)$ assuming distinct eigenvalues with columns ordered by decreasing eigenvalue: $\lambda_1 > \lambda_2 > \lambda_3$. Since the covariance matrix is symmetric, all eigenvalues are real. The canonical coordinates are then:
\begin{equation}
   \phi(V) \myeq  (V-\mathbf{1}\bar{v}^\top)U .
\end{equation}
The eigenvector matrix $U$ is not fully determined because each eigenvector admits a sign flip: if $u$ is an eigenvector, so is $-u$~\cite{ma2024canonicalization}. We resolve this sign ambiguity with a majority rule that orients each axis so that most projected points have positive coordinates.\footnote{The majority rule can fail when the point distribution is exactly symmetric about a principal axis. We acknowledge this as a limitation.} 
For example, let $v_{n,centered}=v_n-\bar{v}$. Then, an eigenvector $u_k$ is determined as
\begin{equation}
     u_k \leftarrow \text{sgn}\left( \sum_n \text{sgn}(v_{n,centered }^{\top} u_k) \right) u_k.
\end{equation}
Algorithm \ref{alg:canonicalization} shows how to canonicalize coordinates with majority-rule sign resolution.

\begin{algorithm}[t]
\caption{Coordinate canonicalization $\phi(\cdot)$ with majority-rule sign resolution}
\label{alg:canonicalization}
\begin{algorithmic}[1]
\REQUIRE Sampled coordinates $V \in \mathbb{R}^{N \times 3}$
\ENSURE Canonical coordinates $\tilde{V} \in \mathbb{R}^{N \times 3}$
\STATE $\bar{v} \gets \frac{1}{N}\sum_{n=1}^{N} v_n$ \hfill $\triangleright$ centroid
\STATE $V_c \gets V - \mathbf{1}\bar{v}^{\top}$ \hfill $\triangleright$ center the point set
\STATE $(\lambda_1,\lambda_2,\lambda_3),\, U \gets \mathrm{eig}(V_c^{\top} V_c)$ \hfill $\triangleright$ columns $u_1,u_2,u_3$ with $\lambda_1 > \lambda_2 > \lambda_3$
\FOR{$k = 1$ \TO $3$}
    \STATE $\mathbf{p} \gets V_c\, u_k$ \hfill $\triangleright$ projection of every point onto axis $k$
    \STATE $n^{+} \gets |\{\, n : p_n > 0 \,\}|$, \quad $n^{-} \gets |\{\, n : p_n < 0 \,\}|$
    \IF{$n^{+} < n^{-}$}
        \STATE $u_k \gets -u_k$ \hfill $\triangleright$ flip so that the majority is positive
    \ENDIF
\ENDFOR
\STATE $\tilde{V} \gets V_c\, U$
\RETURN $\tilde{V}$
\end{algorithmic}
\end{algorithm}

We now show that, once the sign is resolved, the canonical coordinates are the same regardless of how the original point set was oriented.
Before proceeding, we note that any E(3) transformation is represented by a matrix $R \in \mathbb{R}^{3 \times 3}$ with $R^\top R=I$ and a translation vector $t \in \mathbb{R}^{3}$.
Consider the transformed point set $V' \myeq VR^{\top}+\mathbf{1}t^\top$ (i.e., $v'_n = Rv_n + t$). Its centroid is
\begin{equation}
\label{eq:centroid-transform}
    \bar{v}'=\frac{1}{N}\sum_{n=1}^N v_n'=\frac{1}{N}\sum_{n=1}^N (Rv_n + t)=R\bar{v}+t,
\end{equation}
and the centered matrix satisfies
\begin{equation}
\label{eq:cov-transform}
    \begin{split}
            V'-\mathbf{1}\bar{v}'^\top &= VR^{\top}+\mathbf{1}t^\top - \mathbf{1}(R\bar{v}+t)^\top \\
            &=(V-\mathbf{1}\bar{v}^\top)R^\top,
    \end{split}
\end{equation}
so that $\text{cov}(V') = R\,\text{cov}(V)\,R^\top$. Throughout, we assume $\text{cov}(V)$ has distinct eigenvalues.

\begin{lemma}[Sign rule commutes with E(3)]
\label{lem:sign}
If $u$ is a column of the eigenvector matrix $U$ of $\text{cov}(V)$, then $Ru$ is the corresponding column of the eigenvector matrix of $\text{cov}(V')$, and the majority rule (Algorithm~\ref{alg:canonicalization}) assigns $Ru$ the same sign it assigns $u$.
\end{lemma}
\begin{proof}

By \eqref{eq:cov-transform}, $\text{cov}(V')$ and $\text{cov}(V)$ share the same eigenvalues because $R$ is orthogonal, so the ordering $\lambda_1 > \lambda_2 > \lambda_3$ is preserved, and if $U$ is the eigenvector matrix of $\text{cov}(V)$, then $RU$ is the eigenvector matrix of $\text{cov}(V')$.
Since $v_n'-\bar{v}'=R(v_n-\bar{v})$ and $R^{\top}R=I$, we have $(v_n'-\bar{v}')^\top(Ru)=(v_n-\bar{v})^{\top}u$ for every column $u$ of $U$, so the counts $n^{+}$ and $n^{-}$ in Algorithm~\ref{alg:canonicalization} are identical for $u$ and $Ru$, and the majority rule selects the sign of $Ru$ consistently with that of $u$.
\end{proof}

\begin{theorem}[E(3)-invariance of $\phi$]
\label{thm:invariance}
Given $V \in \mathbb{R}^{N\times 3}$ whose covariance matrix has distinct eigenvalues, the canonical coordinates $\tilde{V} \myeq \phi(V)$ are invariant under E(3) transformations, i.e., $\phi(V') = \phi(V)$.
\end{theorem}
\begin{proof}
By Lemma~\ref{lem:sign}, the sign-resolved eigenvector matrices of $\text{cov}(V)$ and $\text{cov}(V')$ are $U$ and $RU$. Using \eqref{eq:cov-transform},
\begin{equation}
    \begin{split}
            \tilde{V}' & \myeq (V'-\mathbf{1}\bar{v}'^\top)RU \\
  &=(V-\mathbf{1}\bar{v}^\top)R^\top RU \\
  &= (V-\mathbf{1}\bar{v}^\top)U \\
   &\myeq \tilde{V} .
    \end{split}
\end{equation}
\end{proof}

Theorem~\ref{thm:invariance} concerns $\phi$ applied to a fixed point set.
The sampler $\mathbf{s}(\cdot)$ is stochastic, so the full pipeline $\phi\circ\mathbf{s}$ is invariant in distribution rather than pointwise.

\subsection{E3PN}\label{sec:e3pn}
Combining the above components with the classifier $\psi_{\theta}$ yields E3PN. 
Finally, we train E3PN with the cross-entropy loss. Given a training set 
$\mathcal{D} = \{(X^{(i)}, y^{(i)})\}$, the loss is
\begin{equation}
    \mathcal{L}(\theta) = -\frac{1}{|\mathcal{D}|}\sum_{i=1}^{|\mathcal{D}|} \sum_{c=1}^{C}     \mathbbm{1} [y^{(i)}=c] \log \hat{y}_c^{(i)},
\end{equation}
where $\hat{y}^{(i)} = \psi_{\theta}(\phi(\mathbf{s}(X^{(i)}))) \in [0,1]^{C}$ 
is the predicted class probability vector (softmax output), $\hat{y}_c^{(i)}$ denotes its $c$-th entry, and $\mathbbm{1}[\cdot]$ is the indicator function.

Now, we discuss the time complexity of E3PN. If we adopt PointNet~\cite{qi2017pointnet} as an efficient classifier $\psi_{\theta}$, then it requires only $O(N \cdot d)$ time for each image consisting of $N$ points and latent dimension $d$. Since the time complexity of the sampling module is $O(M)$ and the coordinate canonicalization module requires $O(N)$ time (note that the eigendecomposition of a $3 \times 3$ matrix can be considered a constant-time operation), the total time complexity is $O(M+N\cdot d)$, where $M$ is the number of non-zero-intensity voxels, $N$ is the number of sampled points, and $d$ is the latent dimension.

\section{Experiments}\label{sec:experiments}

\subsection{Dataset}

We utilize two datasets in our experiments. The first is an ADPKD dataset comprising MR images paired with genetic mutation labels. There are five types of labels: PKD1 truncating, PKD1 non-truncating, PKD2 truncating, PKD2 non-truncating, NMD (no mutation detected).
A total of 478 cases were collected from the University of Pittsburgh.
The experimental protocols were approved by the Institutional Review Board of the university. The need for informed consent was waived by the IRB of the university due to the retrospective/anonymous nature of the study. All methods were carried out in accordance with relevant guidelines and regulations.
After excluding cases in which MR images were acquired using T1-weighted sequences or genetic mutation information was ambiguous, 441 cases were retained to construct the final experimental dataset. Since the original MR images contain regions beyond the kidneys, we applied segmentation masks \cite{kim2022deep} to extract only the kidney regions. The second is ShapeNet, a large-scale 3D shape dataset, which we use to enable a fair and direct comparison with the baseline method. The detailed dataset statistics are summarized in Table~\ref{tab:dataset}.

\begin{table}[!t]
\centering
\begin{tabular}{@{}lcc@{}}
\toprule & \textbf{ADPKD} & \textbf{ShapeNet} \\
\midrule
Data Type                     & Voxel & Point Cloud \\
\# Classes                    & 5 & 16 \\
\# Total Samples              & 441 & 16,846 \\
Resolution                    & $256 \times 256 \times 64$ & -- \\
\bottomrule
\end{tabular}
\caption{Dataset Statistics}
\label{tab:dataset}
\end{table}

\subsection{Experimental settings}
Our model is implemented using PyTorch. Given that the ADPKD dataset is imbalanced and small in scale, we apply 10-fold cross-validation to ensure reliable and unbiased evaluation. The model is trained for 100 epochs with a learning rate of $10^{-3}$ and a batch size of 8. For point cloud-based models, 1,024 points are sampled from each input for use in training and evaluation. The density-based point sampling parameters are estimated using scikit-learn. We also utilize PointNet~\cite{qi2017pointnet} as the classifier $\psi$ with one minor modification: the input dimension is increased from 3 to 4 so that its input layer to accept 4-channel points (three coordinates plus intensity). All experiments are conducted on a workstation equipped with an Intel Xeon Silver 4410Y CPU, 128GB of RAM, and an NVIDIA RTX 6000. 

To the best of our knowledge, no prior work has directly addressed five-way classification of ADPKD genetic mutation variants from 3D MRI.
Existing studies on ADPKD primarily focus on kidney segmentation or total kidney volume estimation rather than genotype prediction, and publicly available benchmarks for this specific task do not exist. We therefore compare E3PN against two representative models as follows.
ResNet~\cite{he2016deep} is a widely adopted convolutional neural network that leverages residual connections to enable effective training of deep architectures, and has been extensively used as a strong baseline in various classification tasks. e3nn~\cite{geiger2022e3nn} is an E(3)-equivariant network implemented with the e3nn library, which processes geometric data while preserving rotational and translational equivariance through irreducible representations. 
For a fair comparison, e3nn receives the same input as E3PN: 1,024 points per image drawn with our density-based sampling. Thus, the two models differ only in how invariance is obtained. ResNet, being a convolutional model, operates directly on a voxel volume. 
By comparing with these two models, we can evaluate E3PN from both a conventional image-based perspective and an equivariant point cloud-based perspective.

\begin{table}[!t]
\centering
\begin{tabular}{@{}lccc@{}}
\toprule
Model & Precision & Recall & F1 score \\
\midrule
ResNet~\cite{he2016deep} & 0.1554 $\pm$ 0.0512 & 0.1652 $\pm$ 0.0187 & 0.1648 $\pm$ 0.0402 \\
e3nn~\cite{geiger2022e3nn}  & 0.1684 $\pm$ 0.0568 & 0.1739 $\pm$ 0.0265 & 0.1633 $\pm$ 0.0307 \\
E3PN  & \underline{0.2538 $\pm$ 0.0627} & \underline{0.2412 $\pm$ 0.0506} & \underline{0.2259 $\pm$ 0.0622} \\
\bottomrule
\end{tabular}
\caption{Performance comparison of ResNet, e3nn, and E3PN on the ADPKD dataset. \underline{Underline} indicates the best performance.}
\label{tab:performance}
\end{table}

\subsection{Comparison results}\label{sec:Comparisonresults}

\begin{figure}[!t]
    \centering
    \begin{subfigure}[b]{0.4\textwidth}
        \centering
        \includegraphics[width=\textwidth]{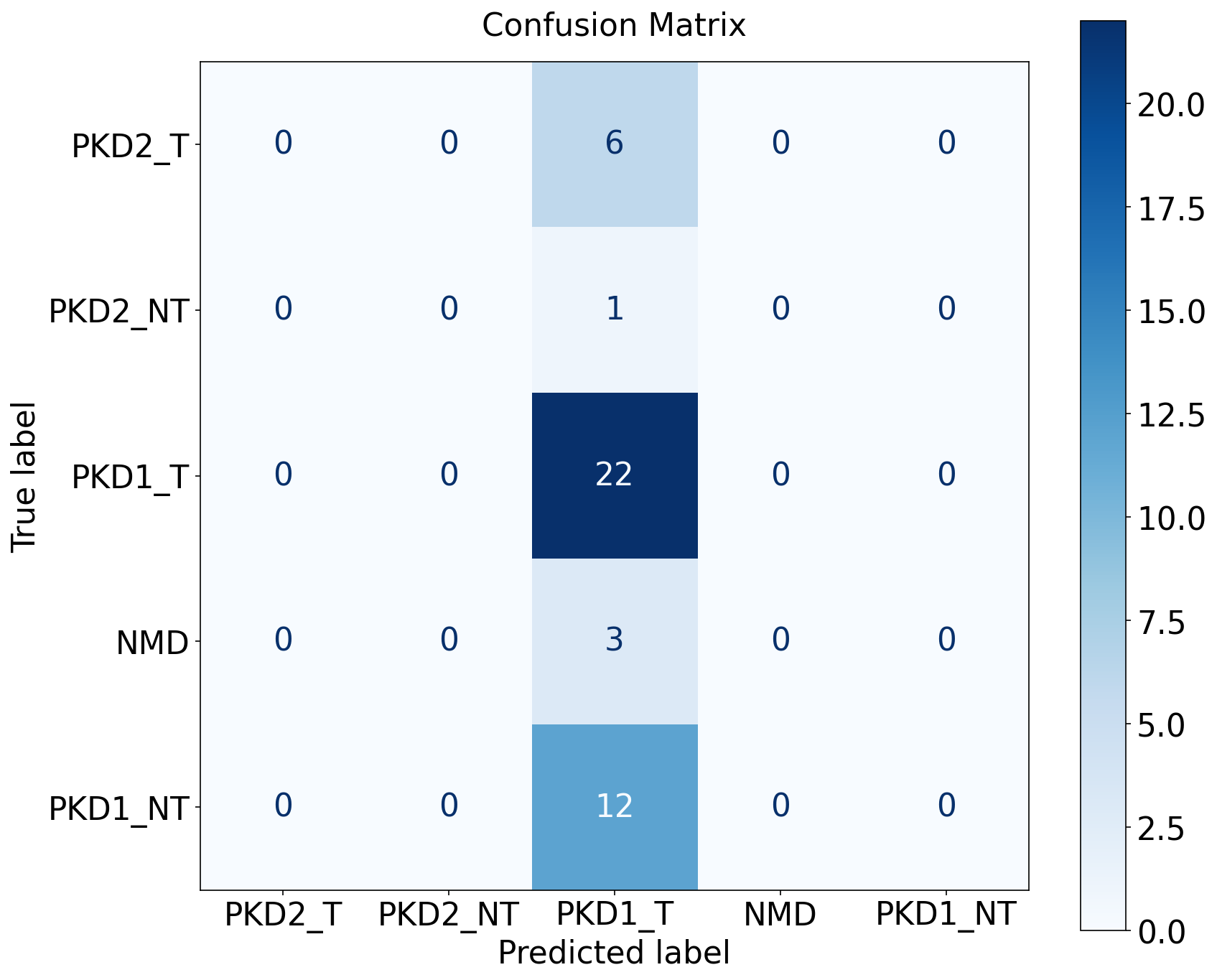}
        \caption{ResNet}
    \end{subfigure}
    \hspace{0.5cm}
    \begin{subfigure}[b]{0.4\textwidth}
        \centering
        \includegraphics[width=\textwidth]{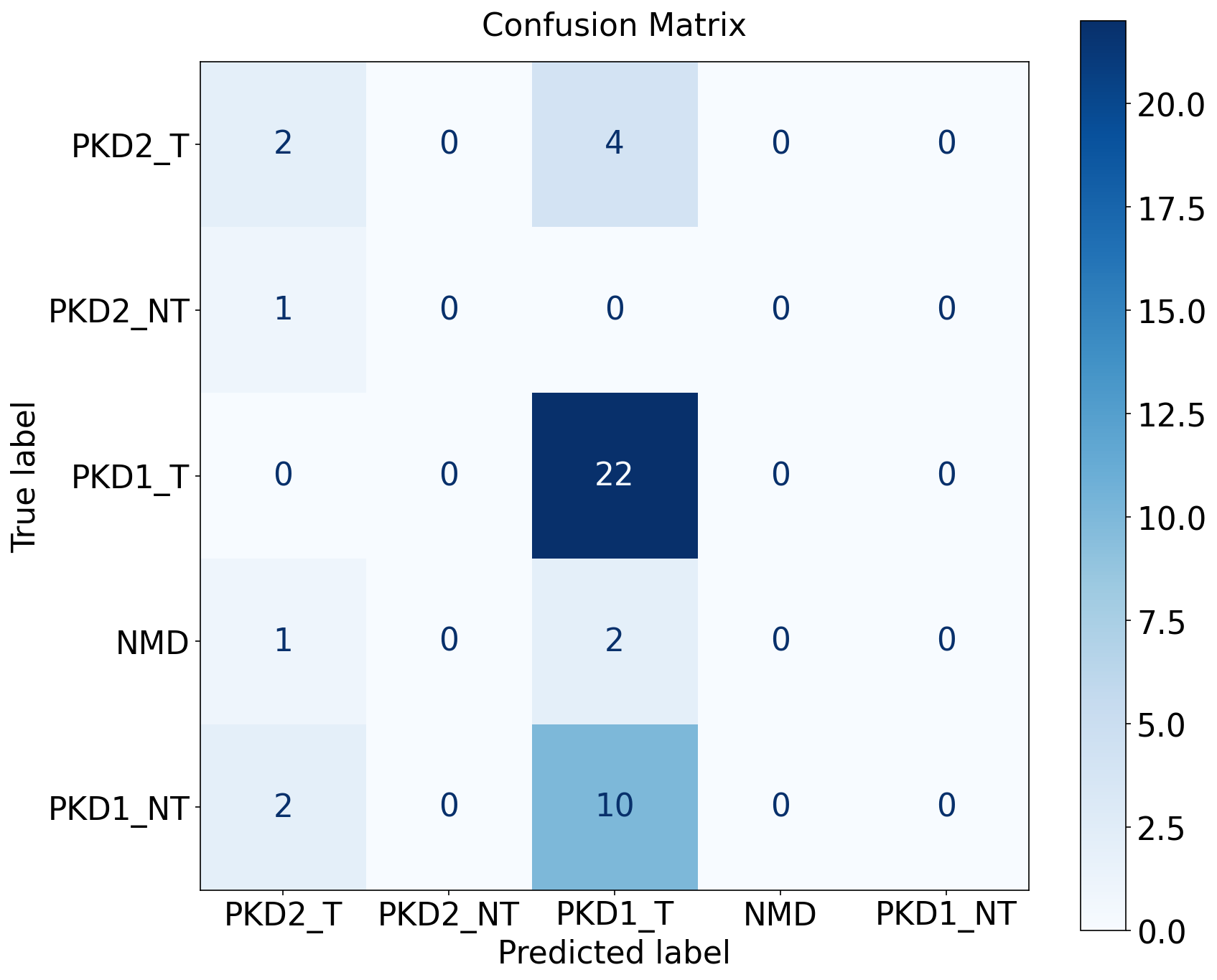}
        \caption{e3nn}
    \end{subfigure}    
    \hspace{0.5cm}
    \begin{subfigure}[b]{0.4\textwidth}
        \centering
        \includegraphics[width=\textwidth]{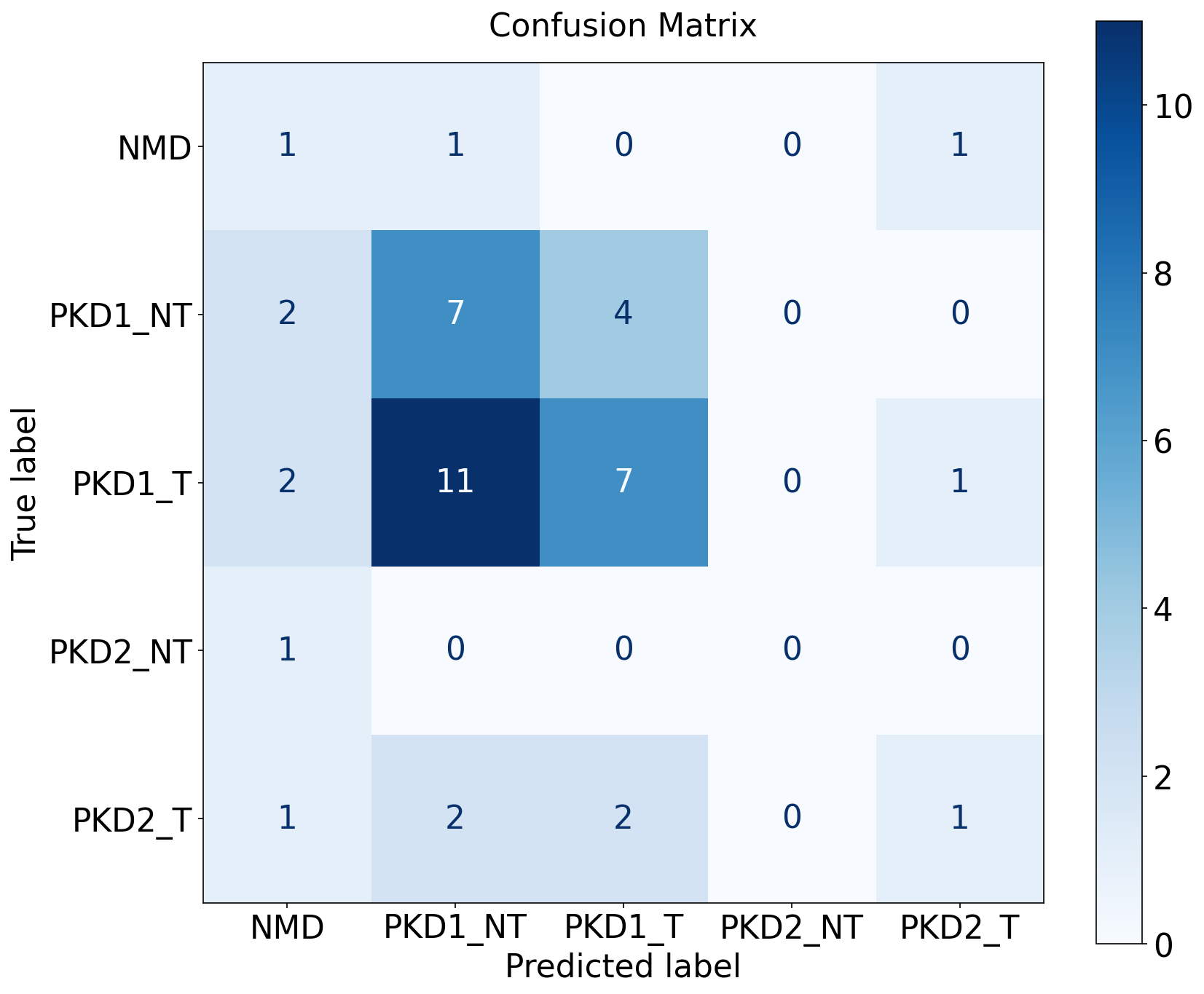}
        \caption{E3PN}
    \end{subfigure}
    \caption{Confusion matrices for (a) ResNet, (b) e3nn, and (c) E3PN on the ADPKD dataset.}
    \label{fig:confusion}
\end{figure}

\begin{table}[!t]
\centering
\begin{tabular}{@{}lccc@{}}
\toprule
Model & Precision & Recall & F1 score \\
\midrule
e3nn  & 0.8672 & 0.8542 & 0.8577 \\
E3PN  & \underline{0.9314} & \underline{0.9660} & \underline{0.9457} \\
\bottomrule
\end{tabular}
\caption{Performance comparison of e3nn and E3PN on the ShapeNet dataset. \underline{Underline} indicates the best performance.}
\label{tab:shapenet_performance}
\end{table}

Table~\ref{tab:performance} should be read against two reference points.
A classifier that guesses randomly in proportion to the class priors attains an expected macro-F1 of 0.20 regardless of class balance, and a uniformly random classifier attains at most 0.20 (lower under imbalance).
A majority-class predictor that always outputs PKD1 truncating attains roughly 0.10--0.15 given the class proportions. 
ResNet and e3nn (F1 = 0.165 and 0.163) sit at or below these references, whereas E3PN reaches 0.226, a modest but consistent improvement across the ten folds.

The confusion matrices in Figure~\ref{fig:confusion} explain the gap. 
Because the cohort is small and heavily imbalanced, both baselines collapse toward the majority class and predict PKD1 truncating regardless of the true label, which depresses their precision and recall. 
E3PN distributes its predictions across more classes, which accounts for its higher scores. 
Figure~\ref{fig:confusion}(c) shows, however, that the improvement is concentrated in the two PKD1 subtypes: 
PKD1 truncating and non-truncating cases are partially separated from each other, whereas PKD2 truncating, PKD2 non-truncating, and NMD cases are rarely predicted correctly and are most often assigned to a PKD1 class. 
The 0.06 absolute (37\% relative) gain in F1 therefore reflects better handling of the two majority classes rather than recovery of the minority genotypes. 
This pattern is consistent with the clinical picture: PKD1 truncating mutations produce the most aggressive morphological phenotype, so the PKD1 distinction is the one most plausibly visible in imaging, while PKD2 and NMD cases are both rarer in the cohort (the two PKD2 classes together comprise only 7 cases in a test fold) and morphologically milder.

We stress that these results should not be taken as evidence that MRI can substitute for genetic testing. Rather, they quantify how much genotype signal a purely geometric model can extract from a 441-patient cohort, and they indicate that larger cohorts or fusion with clinical covariates will be necessary for clinically useful accuracy. 
The ShapeNet results below suggest that the limitation lies in the data regime rather than the architecture.

Table~\ref{tab:shapenet_performance} shows experimental results on ShapeNet, a widely used benchmark dataset in 3D point cloud classification. E3PN improves upon e3nn by 7.40\% in precision, 13.09\% in recall, and 10.26\% in F1 score in relative terms. Notably, the same model reaches an F1 score above 0.94 when training data is plentiful, suggesting that the results on ADPKD stem from factors such as data scarcity rather than architectural limitations.

\begin{table}[!t]
\centering
\footnotesize
\setlength{\tabcolsep}{5pt}
\begin{tabular}{@{}lcccc@{}}
\toprule
\multirow{2}{*}{Model} & \multicolumn{2}{c}{Time / Epoch} & \multicolumn{2}{c}{GPU Memory} \\
\cmidrule(lr){2-3} \cmidrule(lr){4-5}
        & (s)              & Speed-up      & (GB)              & Reduction \\
\midrule
ResNet  & 32.7             & --            & 31.784            & -- \\
e3nn    & 10.7             & 3.1$\times$   & 0.952             & 33.4$\times$ \\
E3PN    & \underline{0.2}  & 163.5$\times$ & \underline{0.307} & 103.5$\times$ \\
\bottomrule
\end{tabular}
\caption{Efficiency comparison of ResNet, e3nn, and E3PN on the ADPKD dataset. \underline{Underline} indicates the best performance.}
\label{tab:efficiency}
\end{table}

\subsection{Efficiency test}

3D voxel grids are large yet mostly empty, so training on them incurs high computational and memory costs.
Furthermore, e3nn relies on spherical harmonics-based irreducible representations and performs tensor product operations at every layer. 
These tensor products compute couplings between features of different orders, and their computational cost grows quickly. 
E3PN bypasses these costs: since canonicalization enforces invariance before the network, no tensor products or irreducible representations are needed, and the classifier consists of inexpensive deep learning models.
Additionally, by converting voxel data to point clouds, the input size is greatly reduced, further contributing to the efficiency gains.
As summarized in Table~\ref{tab:efficiency}, our method is approximately 53 times faster in training time per epoch and requires 3.1 times less GPU memory compared to e3nn, and 163 times faster with 103 times less memory compared to ResNet.

\subsection{Ablation study}
\begin{table}[!t]
\centering
\begin{tabular}{@{}lc@{}}
\toprule
Model Component & F1 score \\
\midrule
w/o coordinate canonicalization                 & 0.1963 $\pm$ 0.0519 \\
w/o density-based point sampling                    & 0.2165 $\pm$ 0.0567 \\
E3PN                           & \underline{0.2259 $\pm$ 0.0622} \\
\bottomrule
\end{tabular}
\caption{Ablation study on the ADPKD dataset. \underline{Underline} indicates the best performance.}
\label{tab:ablation}
\end{table}

To evaluate the contribution of each component in E3PN, we conduct an ablation study. The results are summarized in Table~\ref{tab:ablation}.
In this table, ``w/o coordinate canonicalization'' means without the coordinate canonicalization module so we simply exclude the module. In the ``w/o density-based point sampling'' setting, we use the entire points that have non-zero intensity. 
Removing coordinate canonicalization causes the largest drop in F1 score, while removing density-based sampling also degrades performance.

\begin{figure}[!t]
    \centering
    \begin{subfigure}[b]{0.3\textwidth}
        \centering
        \includegraphics[width=\textwidth]{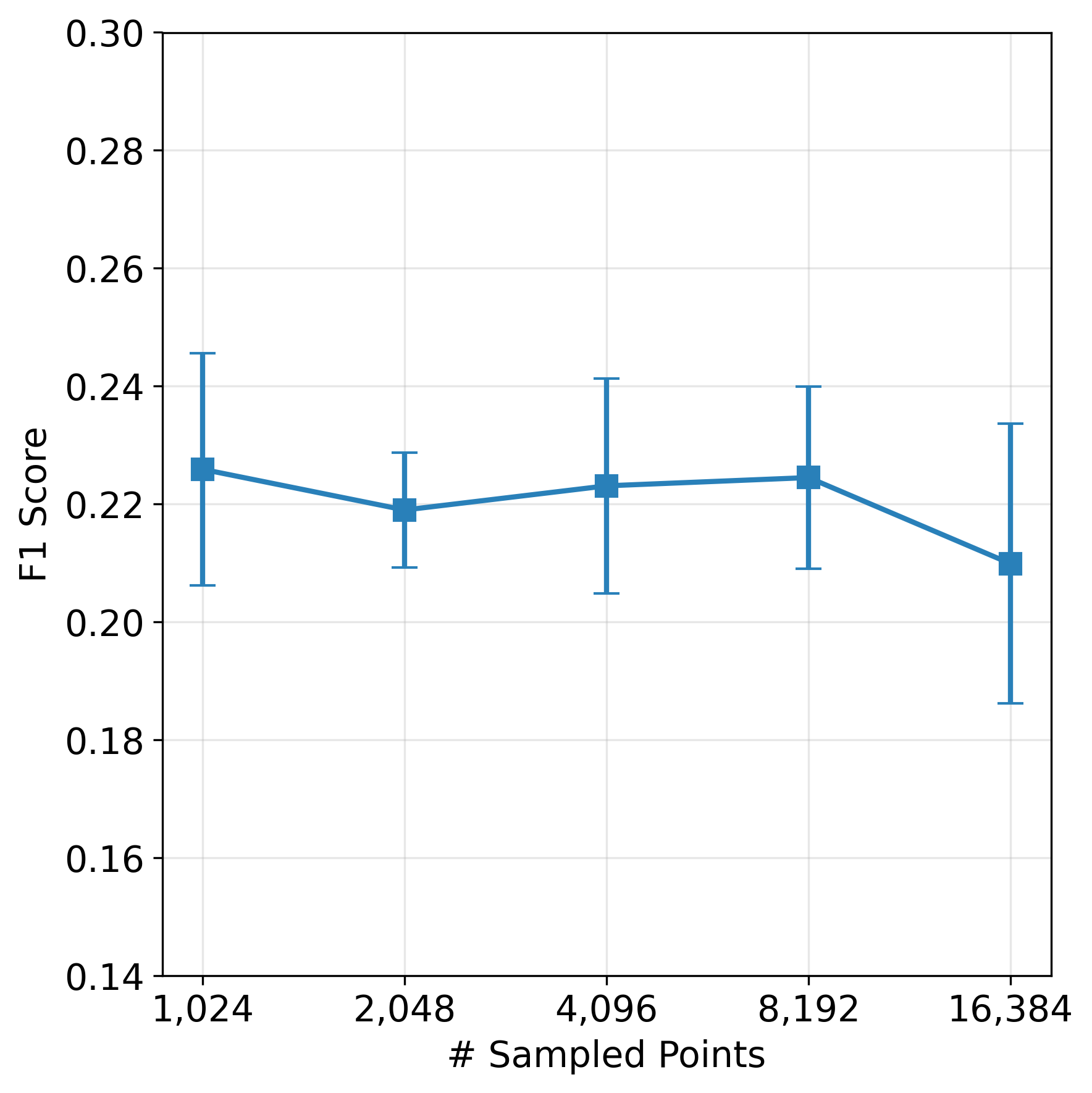}
        \caption{Sampled Points}
    \end{subfigure}
    \hspace{0.5cm}
    \begin{subfigure}[b]{0.3\textwidth}
        \centering
        \includegraphics[width=\textwidth]{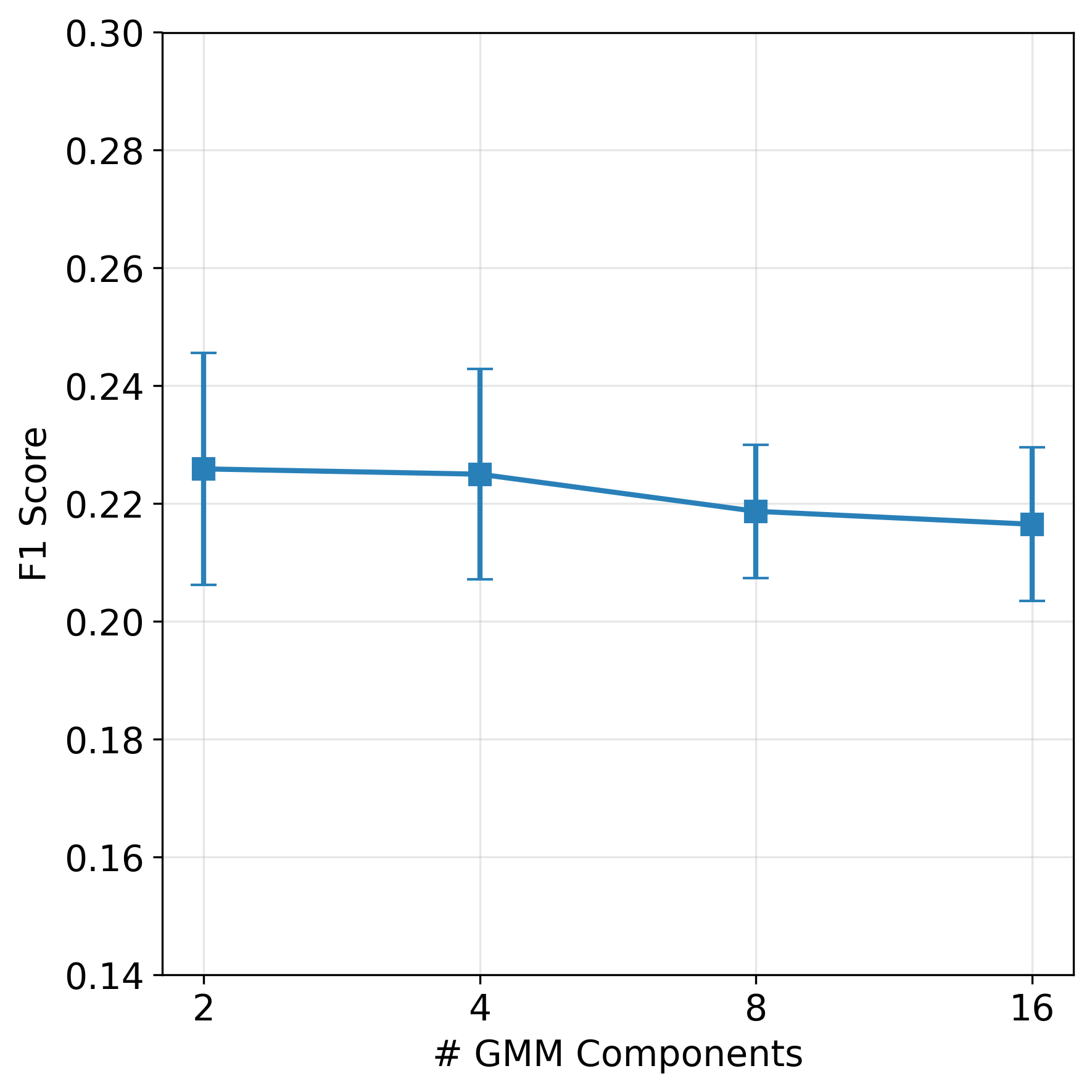}
        \caption{GMM Components}
    \end{subfigure}    
    \hspace{0.5cm}
    \caption{Ablation study of E3PN on the ADPKD dataset. (a) Effect of varying the number of sampled points from the point cloud. (b) Effect of varying the number of GMM components.}
    \label{fig:parameter}
\end{figure}

The ablation results in Figure~\ref{fig:parameter}(a) show that classification performance saturates beyond 1,024 sampled points. Therefore, we adopt 1,024 as the default sampling size to maximize computational efficiency without compromising classification performance. Similarly, as shown in Figure~\ref{fig:parameter}(b), varying the number of GMM components from 2 to 16 shows no substantial difference in F1 score. Accordingly, we use two components as the default setting, as it shows  stable performance while minimizing model complexity.

\section{Conclusion}\label{sec:conclusion}

We presented E3PN, an efficient E(3)-invariant network for classifying ADPKD genotypes from 3D abdominal MRI. 
E3PN represents each kidney as a compact point cloud sampled from a Gaussian-mixture density and removes orientation variability through coordinate canonicalization, addressing both the memory cost of volumetric data and the scarcity of genotype-labeled scans. 
On the five-way genotype task, all methods remain near chance, and we report this as a central finding: genotype is not readily recoverable from kidney geometry alone at current data scales. Within that regime, E3PN trains in 0.2s per epoch on a single GPU and yields the best F1 of the three methods, making it a practical baseline for future work that fuses imaging with clinical covariates or scales to larger cohorts.

\section*{ACKNOWLEDGMENT}
This work was supported by the National Research Foundation of Korea (NRF) grant
funded by the Korea government (MSIT)(RS-2023-00278009). This work was also supported by the National Research Foundation of Korea (NRF) grant funded by the Korea government (MSIT) (No. RS-2024-00345567)

\section*{Declaration of generative AI and AI-assisted technologies in the proofreading}
During the preparation of this work the authors used ChatGPT and Claude in order to improve language and readability, with caution. After using this service, the authors reviewed and edited the content as needed and take full responsibility for the content of the publication.

\bibliographystyle{JISEbib}
\bibliography{ref}

\end{document}